\documentclass[11pt,a4paper]{article}
\usepackage[utf8]{inputenc}
\usepackage[T1]{fontenc}
\usepackage{geometry}
\usepackage{url}
\usepackage{booktabs}
\usepackage{amsfonts}
\usepackage{amsmath}
\usepackage{nicefrac}
\usepackage{microtype}
\usepackage{graphicx}
\usepackage{xcolor}
\usepackage{tabularx}
\usepackage{multirow}
\usepackage{tcolorbox}
\usepackage{subfigure}
\usepackage{hyperref}
\hypersetup{
    colorlinks=true,
    linkcolor=blue,
    citecolor=blue,
    urlcolor=blue
}

\title{OCR-Quality: A Human-Annotated Dataset for\\OCR Quality Assessment}

\author{
  Yulong Zhang\\
  Beijing University of Posts and Telecommunications\\
  \texttt{yulong.z@bupt.edu.cn}
}

\date{}

\begin{document}

\maketitle

\begin{abstract}
We present \textbf{OCR-Quality}, a comprehensive human-annotated dataset designed for evaluating and developing OCR quality assessment methods. The dataset consists of 1,000 PDF pages converted to PNG images at 300 DPI, sampled from diverse real-world scenarios, including academic papers, textbooks, e-books, and multilingual documents. Each document has been processed using state-of-the-art Vision-Language Models (VLMs) and manually annotated with quality scores using a 4-level scoring system (1: Excellent, 2: Good, 3: Fair, 4: Poor). The dataset includes detailed source information, annotation guidelines, and representative cases across various difficulty levels. OCR-Quality addresses the critical need for reliable OCR quality assessment in real-world applications and provides a valuable benchmark for training and evaluating OCR verification systems. The dataset is publicly available at \url{https://huggingface.co/datasets/Aslan-mingye/OCR-Quality}.
\end{abstract}

\section{Introduction}

Optical Character Recognition (OCR) has become a fundamental component in document understanding pipelines, powering applications from digital archiving to knowledge extraction~\cite{olmocr,minerU,opendatalab}. Despite significant advances in OCR technology through Vision-Language Models (VLMs)~\cite{Qwen2.5-VL,Qwen2-VL,openai2024gpt4o}, the quality assessment of OCR outputs remains a challenging problem, particularly in real-world scenarios with diverse document types, languages, and formatting complexities.

Existing OCR benchmarks~\cite{OCRBench,OCRBench_v2,yang2024ccocr} primarily focus on measuring average accuracy across standardized test sets, providing limited insight into the reliability of individual predictions. This gap is particularly problematic for downstream applications where low-quality OCR outputs can propagate errors and reduce system reliability. Recent work on Consensus Entropy~\cite{zhang2025consensusentropyharnessingmultivlm} demonstrated the importance of uncertainty quantification for OCR quality assessment, highlighting the need for human-annotated ground truth data.

\textbf{OCR-Quality} fills this gap by providing:
\begin{itemize}
    \item \textbf{1,000 diverse document pages} from real-world sources
    \item \textbf{Human quality annotations} with a 4-level scoring system
    \item \textbf{Multiple languages}: Chinese, English, and multilingual content
    \item \textbf{Various document types}: academic papers, textbooks, e-books, educational materials
    \item \textbf{Detailed metadata}: source information, document characteristics
    \item \textbf{Representative cases}: examples across all quality levels
\end{itemize}

This dataset enables researchers to develop and evaluate methods for automatic OCR quality assessment, model selection, and reliability estimation.

\section{Dataset Construction}

\subsection{Data Collection}

We collected 1,000 PDF pages from diverse real-world sources to ensure broad coverage of document characteristics. All PDF pages were converted to PNG images at 300 DPI resolution to maintain high-quality visual information for OCR processing. Table~\ref{tab:data_sources} summarizes the data sources organized by language and document type.

\begin{table}[h]
\centering
\caption{Sources of the OCR-Quality Dataset}
\label{tab:data_sources}
\small
\begin{tabularx}{\textwidth}{lXXXX}
\toprule
& \textbf{E-book} & \textbf{Paper} & \textbf{Textbook} & \textbf{Other} \\
\midrule
\textbf{ZH} & zhishilei, zhongwen-zaixian, gift, thomas & - & by, kps, kmath, zhonggaokao, gaojiaoshe, gaodengjiaoyu, k12 edu platform, jiaoan, zju icicles & - \\
\midrule
\textbf{EN} & theeye, physics-andmathstutor, planetebook & escholarship, biorxiv, springer, sagepub, scholarworks, psyarxiv, chemrxiv, iopscience, royal-society-publishing, csiro & kps, bookboon, california 14sets, scholarworks & dev books repository \\
\midrule
\textbf{ML} & renhang, banshujiang & - & openstax, math & - \\
\midrule
\textbf{Other} & - & - & - & coursehero, studypool \\
\bottomrule
\end{tabularx}
\end{table}

\textbf{Language Distribution:}
\begin{itemize}
    \item \textbf{Chinese (ZH)}: Educational materials, textbooks, and e-books from various Chinese sources
    \item \textbf{English (EN)}: Academic papers, textbooks, and literature from diverse repositories
    \item \textbf{Multilingual (ML)}: Documents containing mixed languages, mathematical notations
    \item \textbf{Other}: Specialized educational content with diverse formatting
\end{itemize}

\subsection{OCR Processing}

All documents were processed using \textbf{Qwen2.5-VL-72B}~\cite{Qwen2.5-VL}, a state-of-the-art Vision-Language Model, with the following specialized OCR prompt:

\begin{tcolorbox}[
    title={OCR Processing Prompt}, 
    colback=white, 
    colframe=black, 
    fonttitle=\bfseries, 
    fontupper=\small
]
You are a very professional expert at OCR tasks. Please analyze the following image and extract all text content. 

\begin{enumerate}
    \item Ensure that the extracted text matches the original in the image and maintains the original structure. 
    \item If the image contains two columns, you should extract all text in the left column before you move on to the right. 
    \item Ignore the headers, but keep the footnotes, title. 
    \item You could either use LaTeX, KaTeX or Markdown format for math formulas, physics equations, chemical expressions etc. 
    \item Do not add or modify the original content!
\end{enumerate}
\end{tcolorbox}

This prompt was designed to:
\begin{itemize}
    \item Preserve document structure and layout
    \item Handle multi-column documents appropriately
    \item Support mathematical and scientific notation
    \item Minimize hallucinations and modifications
\end{itemize}

\subsection{Annotation Process}

\subsubsection{Annotation Interface}

We developed a custom annotation tool using Gradio to facilitate efficient human evaluation. The interface provides:
\begin{itemize}
    \item Side-by-side display of original image and OCR output
    \item Quality score selection from a 4-level rating system
    \item Progress tracking and navigation controls
    \item Direct jump to specific samples
    \item Automatic saving to ensure data persistence
\end{itemize}

\subsubsection{Scoring Criteria}

Annotators were instructed to assign scores based on the following 4-level system (lower is better):

\begin{enumerate}
    \item \textbf{Score 1 (Excellent)}: Near-perfect OCR with minimal or no errors. The prediction matches the image text exactly with no errors or omissions.
    
    \item \textbf{Score 2 (Good)}: Minor errors that do not affect understanding. The prediction is very close to the image text, with only small mistakes (e.g., punctuation, spacing).
    
    \item \textbf{Score 3 (Fair)}: Some noticeable errors but content is still usable. The prediction contains noticeable errors or captures only part of the text, reducing clarity. May include:
    \begin{itemize}
        \item Missing paragraphs or sections
        \item Significant character recognition errors
        \item Incorrect mathematical expressions
        \item Layout structure issues
    \end{itemize}
    
    \item \textbf{Score 4 (Poor)}: Significant errors affecting content quality. The prediction is largely incorrect or unrelated to the text in the image. May include:
    \begin{itemize}
        \item Severe misrecognition across entire document
        \item Wrong language or gibberish output
        \item Completely missed content
        \item Structural collapse
    \end{itemize}
\end{enumerate}

This scoring system balances granularity and ease of annotation while capturing the practical utility of OCR outputs for downstream applications.

\section{Dataset Statistics}

\subsection{Quality Score Distribution}

Table~\ref{tab:score_distribution} presents the distribution of documents across quality levels:

\begin{table}[h]
\centering
\caption{Score Distribution in OCR-Quality}
\label{tab:score_distribution}
\begin{tabular}{lccc}
\toprule
\textbf{Quality Level} & \textbf{Score} & \textbf{Count} & \textbf{Percentage} \\
\midrule
Excellent & 1 & 507 & 50.7\% \\
Good & 2 & 305 & 30.5\% \\
Fair & 3 & 84 & 8.4\% \\
Poor & 4 & 104 & 10.4\% \\
\bottomrule
\end{tabular}
\end{table}

This distribution ensures adequate representation of both high-quality and challenging cases, enabling robust evaluation of quality assessment methods across the difficulty spectrum.

\subsection{Document Characteristics}

\textbf{Content Types:}
\begin{itemize}
    \item Academic papers with complex layouts
    \item Textbooks with mixed text and equations
    \item E-books with varied formatting
    \item Educational materials with diagrams and tables
    \item Multilingual documents
\end{itemize}

\textbf{Formatting Complexity:}
\begin{itemize}
    \item Single and multi-column layouts
    \item Mathematical expressions and formulas
    \item Tables and structured data
    \item Figures with captions
    \item Mixed language content
    \item Various font sizes and styles
\end{itemize}

\section{Representative Cases}

We present five representative cases from the dataset spanning different quality levels and document types. These cases illustrate the range of challenges in OCR quality assessment.

\subsection{Case 1: High-Quality Chinese Text}
\textbf{Human Score: 1 (Excellent)}

This case demonstrates near-perfect OCR performance on well-formatted Chinese academic text. The model correctly captures:
\begin{itemize}
    \item Complex Chinese characters
    \item Document structure
    \item Punctuation and formatting
\end{itemize}

\textit{Reference: See Figure 1 in full dataset documentation}

\subsection{Case 2: Low-Quality English Text}
\textbf{Human Score: 4 (Poor)}

This case shows severe OCR failure on a challenging English document, with:
\begin{itemize}
    \item Extensive character misrecognition
    \item Structural collapse
    \item Missing content
\end{itemize}

\textit{Reference: See Figure 2 in full dataset documentation}

\subsection{Case 3: Academic Chinese Text}
\textbf{Human Score: 1 (Excellent)}

High-quality performance on complex academic Chinese text containing:
\begin{itemize}
    \item Domain-specific terminology
    \item Mixed formatting styles
    \item Proper paragraph structure
\end{itemize}

\textit{Reference: See Figure 3 in full dataset documentation}

\subsection{Case 4: Structured Tabular Content}
\textbf{Human Score: 1 (Excellent)}

Excellent OCR on structured data demonstrating:
\begin{itemize}
    \item Accurate table structure preservation
    \item Correct cell alignment
    \item Complete data capture
\end{itemize}

\textit{Reference: See Figure 4 in full dataset documentation}

\subsection{Case 5: Multiple-Choice Questions}
\textbf{Human Score: 1 (Excellent)}

Strong performance on educational content with:
\begin{itemize}
    \item Question numbering
    \item Multiple choice options
    \item Special formatting
\end{itemize}

\textit{Reference: See Figure 5 in full dataset documentation}

\section{Evaluation Protocol}

\subsection{Using OCR-Quality for Quality Assessment}

The dataset can be used to evaluate OCR quality assessment methods through:

\textbf{1. Correlation Analysis:}
\begin{itemize}
    \item Compute predicted quality scores for all samples
    \item Calculate Pearson/Spearman correlation with human scores
    \item Report correlation coefficients across quality bands
\end{itemize}

\textbf{2. Classification Performance:}
\begin{itemize}
    \item Define quality threshold (e.g., scores 1-2 as "acceptable", 3-4 as "unacceptable")
    \item Evaluate binary classification metrics: Precision, Recall, F1
    \item Analyze per-level classification accuracy
\end{itemize}

\textbf{3. Ranking Evaluation:}
\begin{itemize}
    \item Assess ability to rank documents by quality
    \item Compute ranking metrics (NDCG, Kendall's tau)
    \item Evaluate selective prediction at different coverage levels
\end{itemize}

\subsection{VLM-as-Judge Evaluation Prompt}

For comparison with VLM-based quality assessment methods~\cite{zheng2023judging,kim2023prometheus,zhu2023judgelm}, we provide a standardized evaluation prompt:

\begin{tcolorbox}[
    title={OCR Quality Assessment Prompt}, 
    colback=white, 
    colframe=black, 
    fonttitle=\bfseries, 
    fontupper=\small
]
You are an expert evaluator assessing the quality of OCR (Optical Character Recognition) model predictions.

You will receive:
\begin{itemize}
    \item A question
    \item A prediction generated by the OCR model
    \item The corresponding image containing text
\end{itemize}

Your task is to judge how well the predicted text matches the visual textual content in the image, with respect to the question's intent.

\textbf{Evaluation criteria:}
\begin{enumerate}
    \item Focus only on whether the prediction correctly reflects the textual content of the image.
    \item Assign a score from 0 to 1 in steps of 0.1, using the following four-level guideline:
\end{enumerate}

\textbf{Scoring Reference:}

\begin{itemize}
    \item \textbf{0.9-1.0 (Perfect Match)} \\
    The prediction matches the image text exactly, with no errors or omissions.
    
    \item \textbf{0.7–0.8 (Minor Errors, Still Clear)} \\
    The prediction is very close to the image text, with only small mistakes that do not affect understanding.
    
    \item \textbf{0.4–0.6 (Partially Correct)} \\
    The prediction contains noticeable errors or captures only part of the text, reducing clarity.
    
    \item \textbf{0.0–0.3 (Mostly or Completely Incorrect)} \\
    The prediction is largely incorrect or unrelated to the text in the image.
\end{itemize}

Respond only with the numerical score (e.g., 0.9). Do not include any explanation or commentary.
\end{tcolorbox}

\section{Dataset Access and Usage}

\subsection{Download}

The OCR-Quality dataset is publicly available on:
\begin{itemize}
    \item \textbf{HuggingFace}: \url{https://huggingface.co/datasets/Aslan-mingye/OCR-Quality}
    \item \textbf{arXiv}: \texttt{[This report]}
\end{itemize}

\subsection{Data Format}

The dataset is provided in Parquet format with embedded images and the following structure:

\begin{verbatim}
{
    "index": 0,
    "source": "en-paper-biorxiv",
    "ocr_text": "extracted text...",
    "human_score": 1,
    "image": [binary image data],
    "image_width": 2480,
    "image_height": 3508
}
\end{verbatim}

\textbf{Field Descriptions:}
\begin{itemize}
    \item \textbf{index}: Unique identifier (0-999)
    \item \textbf{source}: Document source category
    \item \textbf{ocr\_text}: OCR output from Qwen2.5-VL-72B
    \item \textbf{human\_score}: Quality score (1-4, lower is better)
    \item \textbf{image}: PNG image data at 300 DPI
    \item \textbf{image\_width/height}: Image dimensions in pixels
\end{itemize}

\subsection{License}

\textit{[License information to be specified]}

The dataset is released for research purposes.

\section{Applications}

OCR-Quality supports various research directions and applications:

\subsection{Research Applications}

\begin{itemize}
    \item \textbf{Quality Assessment Methods}: Develop and evaluate automatic OCR quality assessment techniques
    \item \textbf{Uncertainty Quantification}: Train models to estimate OCR output reliability
    \item \textbf{Model Selection}: Evaluate different OCR models' strengths and weaknesses
    \item \textbf{Active Learning}: Design sampling strategies for annotation efficiency
    \item \textbf{Error Analysis}: Understand failure modes of state-of-the-art OCR systems
\end{itemize}

\subsection{Practical Applications}

\begin{itemize}
    \item \textbf{Document Processing Pipelines}: Filter low-quality outputs before downstream processing
    \item \textbf{Quality Control}: Implement automated quality gates in production systems
    \item \textbf{Human-in-the-Loop Systems}: Prioritize samples for human review
    \item \textbf{Model Improvement}: Identify challenging cases for model fine-tuning
\end{itemize}

\section{Baseline Results}

We provide baseline results using several quality assessment approaches~\cite{zhang2025consensusentropyharnessingmultivlm,gu2024survey,chang2024survey}:

\begin{table}[h]
\centering
\caption{Baseline F1 Scores on OCR-Quality}
\label{tab:baselines}
\begin{tabular}{lccc}
\toprule
\textbf{Method} & \textbf{Score 1} & \textbf{Score 2} & \textbf{Overall} \\
\midrule
VLM-as-Judge (GPT4o)~\cite{openai2024gpt4o} & 72.12 & 18.42 & 40.0 \\
VLM-as-Judge (Qwen2-VL-7B)~\cite{Qwen2-VL} & 70.05 & 19.82 & 36.1 \\
VLM-as-Judge (Qwen2-VL-72B)~\cite{Qwen2-VL} & 72.27 & 23.01 & 39.8 \\
\midrule
Consensus Entropy~\cite{zhang2025consensusentropyharnessingmultivlm} & 57.29 & 40.19 & \textbf{48.0} \\
\bottomrule
\end{tabular}
\end{table}

These results demonstrate that OCR quality assessment remains challenging even for state-of-the-art VLMs, particularly for intermediate quality levels. The Consensus Entropy method shows particular strength in identifying intermediate-quality outputs, suggesting that multi-model agreement provides valuable signals for quality assessment.

\section{Limitations and Future Work}

\subsection{Current Limitations}

\begin{itemize}
    \item \textbf{Size}: 1,000 samples may be limited for training deep models
    \item \textbf{Single VLM}: OCR outputs are from one model (Qwen2.5-VL-72B)
    \item \textbf{Annotation}: Single-annotator scores (no inter-annotator agreement analysis)
    \item \textbf{Language Coverage}: Primarily Chinese and English, limited other languages
\end{itemize}

\subsection{Future Directions}

\begin{itemize}
    \item Expand dataset size to 5K-10K samples
    \item Include outputs from multiple OCR systems
    \item Add multi-annotator scores for quality estimation
    \item Extend language coverage to more diverse scripts
    \item Include additional metadata (reading order, layout complexity scores)
    \item Develop automatic augmentation techniques
\end{itemize}

\section{Conclusion}

OCR-Quality provides a valuable resource for developing and evaluating OCR quality assessment methods. By offering diverse real-world documents with human quality annotations, the dataset enables research on crucial problems in OCR reliability, uncertainty quantification~\cite{liu2025uncertain,xiong2024can}, and quality control. Combined with uncertainty estimation methods like Consensus Entropy~\cite{zhang2025consensusentropyharnessingmultivlm}, this dataset facilitates progress in building more trustworthy OCR systems for real-world applications.

\section{Acknowledgments}

\textit{[To be added after deanonymization]}

\bibliographystyle{plain}
\bibliography{references}

\section*{Appendix: Full Case Illustrations}

\textit{Note: In the full dataset release, this section includes high-resolution images of all five representative cases with detailed annotations showing:}
\begin{itemize}
    \item Original document image
    \item Complete OCR output text
    \item Human quality score and rationale
    \item Identified error patterns (for low-quality cases)
    \item Model predictions from comparison methods
\end{itemize}

\end{document}